\documentclass{article}
\usepackage{ijcai24}

\usepackage{times}
\usepackage{soul}
\usepackage{url}
\usepackage[hidelinks]{hyperref}
\usepackage[utf8]{inputenc}
\usepackage[small]{caption}
\usepackage{graphicx}
\usepackage{amsmath}
\usepackage{amsthm}
\usepackage{booktabs}
\usepackage{algorithm}
\usepackage[noend]{algorithmic}
\usepackage[switch]{lineno}
\usepackage[capitalise, noabbrev]{cleveref}

\title{Spend More to Save More (SM$^2$): An Energy-Aware Implementation of Successive Halving for Sustainable Hyperparameter Optimization}

\author{
Daniel Geißler$^1$
\and
Bo Zhou$^{1,2}$\and
Sungho Suh$^{1,2}$\And
Paul Lukowicz$^{1,2}$\\
\affiliations
$^1$German Research Center for Artificial Intelligence (DFKI), Kaiserslautern, Germany\\
$^2$University of Kaiserslautern-Landau (RPTU), Kaiserslautern, Germany\\
\emails
daniel.geissler@dfki.de
}

\begin{document}

\maketitle

\begin{abstract}
A fundamental step in the development of machine learning models commonly involves the tuning of hyperparameters, often leading to multiple model training runs to work out the best-performing configuration.
As machine learning tasks and models grow in complexity, there is an escalating need for solutions that not only improve performance but also address sustainability concerns.
Existing strategies predominantly focus on maximizing the performance of the model without considering energy efficiency.
To bridge this gap, in this paper, we introduce Spend More to Save More (SM$^2$), an energy-aware hyperparameter optimization implementation based on the widely adopted successive halving algorithm. 
Unlike conventional approaches including energy-intensive testing of individual hyperparameter configurations, SM$^2$ employs exploratory pretraining to identify inefficient configurations with minimal energy expenditure.
Incorporating hardware characteristics and real-time energy consumption tracking, SM$^2$ identifies an optimal configuration that not only maximizes the performance of the model but also enables energy-efficient training.
Experimental validations across various datasets, models, and hardware setups confirm the efficacy of SM$^2$ to prevent the waste of energy during the training of hyperparameter configurations.
\end{abstract}

\section{Introduction}
The rising complexity of Artificial Intelligence (AI) applications solved through advanced deep-learning models continuously increases the energy demand of the whole AI sector.
It is estimated that large language models like the popular \mbox{GPT-3} may require up to 1300 MWh of energy for training, whereas inference may account for even more energy consumption over time \cite{patterson2021carbon}. 
This trend is driven by the ubiquity of machine learning applications being present in the majority of our daily lives.
Considering the life cycle of an AI model from initial development towards its final deployment possibilities, there are multiple stages involved that may be optimized in terms of energy efficiency simply due to the waste of resources.

This work accounts for the initial stages of the life cycle, with a focus on optimizing the training process of machine learning models to generate a more energy-aware and efficient solution for AI developers.
Besides the quality of the dataset and model architecture fitment to the desired task, hyperparameter optimization (HPO) is an integral part of a well-suited model.
Two key hyperparameters, namely batch size and learning rate, significantly affect training convergence quality and speed \cite{he2019control}. However, determining the optimal values for these parameters, often hardware and task-specific, remains a challenging task requiring selection by experienced developers at the first attempt.
Various strategies have been developed to improve the HPO problem, yet their only aim is to maximize the model's prediction performance. 
Additionally, due to trends of supporting AI developers' work with high-performance data centers, usually, due to economic reasons, it is a simple but energy-intensive process to spawn multiple training runs to manually explore the hyperparameter space.
While more efficient hardware has continuously been updated to alleviate the training cost of trial and error, there lacks a holistic method that incorporates hardware energy footprint into the model training HPO process.

To resolve this issue, this paper presents Spend More to Save More (SM$^2$), a novel approach combining HPO and energy consumption tracking to generate a profound strategy to sustainably improve the machine learning training process. 
To the best of our knowledge, this work is the first work to optimize the hyperparameters while considering energy consumption.
Our work makes significant contributions to the field of sustainable HPO and can be summarized as follows:
\begin{itemize}
    \item Implementing energy-aware training through hardware-based power monitoring.
    \item Deploying a sequential Successive Halving Algorithm strategy to minimize energy waste. 
    \item Extending the traditional training regime with exploratory components to energy-efficiently explore hyperparameter configurations.
    \item Evaluating the SM$^2$ approach across three different scenarios of models and datasets to prove the possibility of equal model performance while improving energy efficiency.    
\end{itemize}

\section{Related Works}
\subsection{Hyperparameter Optimization}
\label{hyp_opt}
The landscape of HPO covers multiple different methodologies from different areas to explore and manifest the best-performing hyperparameter setting.
In many cases, there is no clear identification of the best-performing strategy possible because the efficiency of the algorithm depends on the respective machine learning problem to solve and the user's preferences.
Nevertheless, there is a trend of different HPO categories being fused to elevate the performance compared to the traditional algorithms. \cite{bischl2107hyperparameter}

Starting with the classic methods, grid search is the most common and simple strategy next to the manual exploration of hyperparameters.
The strategy is based on an initialized grid that covers the range of each hyperparameter \cite{lecun1998gradient}.
Due to its simplicity, it is commonly used but suffers from inefficiencies in high-dimensional scenarios due to many low-performing training runs and their independence from each other.
Another simple but more efficient strategy in this scope is random search \cite{bergstra2012random}. 
Instead of exploring a fixed grid, the hyperparameter space is randomly sampled to provide a more stochastic exploration.
Due to its ease of implementation, it still serves as a benchmark to compare it to more advanced strategies.

Another category involves evolutionary strategies to work out the best hyperparameter setting \cite{beyer2002evolution}.
Such algorithms explore the hyperparameter space based on the biological concept of evolution.
Throughout a fitness function, a population of hyperparameter configurations is evaluated whereas the worst-performing configurations are removed.
Instead of proceeding with the most promising configurations, similar to natural evolution, the next iteration consists of crossovers and mutations of the previous configurations.
Works like \cite{awad2021dehb} and \cite{bochinski2017hyper} confirm the usability of evolutionary strategies as state-of-the-art.

Bayesian optimization constitutes another strategy for HPO, introduced through works like \cite{snoek2012practical}.
Such algorithms can efficiently explore the hyperparameter space by constructing probabilistic surrogate models and using acquisition functions to guide the search.
Introduced by \cite{bergstra2011algorithms}, Tree-structured Parzen Estimators (TPE) is a Bayesian optimization variant that ranks the performance of HPO configurations. 
TPE has shown impressive results, especially in high-dimensional optimization tasks, making it one of the current state-of-the-art HPO algorithms

Hyperparameter optimization through Reinforcement Learning (RL) was recently introduced through works like \cite{jomaa2019hyp} and \cite{wu2023hyperparameter}.
Instead of systematically exploring the hyperparameter space, RL consists of a sequential decision-making process to find the best configuration based on the policy that guides the RL agent along the desired path.
In terms of sustainably and efficiency, this solution is quite complex and requires many training iterations due to the expensive exploration and exploitation trade-off in RL.
Moreover, instead of optimizing only the traditional model, also the RL model requires optimization.

Pruning-based methods, such as the common Successive Halving Algorithm (SHA) \cite{li2018hyperband}, aim to efficiently allocate computational resources to promising hyperparameter configurations in a parallelized environment. 
By iteratively terminating less promising configurations, such methods reduce the overall training time while maintaining a focus on high-performing settings.
This bandit-based algorithm demonstrates a practical approach to balancing exploration and exploitation, making it well-suited for high-dimensional HPO scenarios.
Throughout extended works like \cite{li2020system}, multi-fidelity optimization improves SHA by removing predefined evaluation timestamps.
Instead of training all configurations in a synchronized, parallel environment, resources can be reallocated asynchronously to generate a more dynamic training procedure.
Works like \cite{lee2023improving} and \cite{wistuba2022supervising} are currently considered state-of-the-art in this field.
Due to its benefits in terms of efficiency and its overlaps with our ideas to stop less-promising configurations early in the training process, we utilize the SHA approach as the foundation of SM$^2$.

\subsection{Energy Consumption Tracking}
\label{energy_sec}

In the AI sector, energy consumption tools are still a niche, especially when it comes to generating awareness about the energy expenditure for training machine learning models.
Currently, there is a small but growing list of software available with most of them utilizing the same approach to gather the power consumption data from the hardware manufacturers’ utility logging. 
They are usually designed to capture energy information by building an additional layer between the system’s hardware configuration and the user’s model training process \cite{henderson2020towards}. 
Generally speaking, the power consumption of the Graphics Processing Units (GPU) is considered the largest part of the training process as it performs the core work with the parallel processing of mathematical tasks \cite{mittal2014survey}. 
Nevertheless, other hardware components and even secondary power consumers like the cooling system or the power supply unit (PSU) itself contribute to the overall energy consumption.
Therefore, there is great interest in tracking the power consumption of the full system to minimize deviations.

As a general rule, the energy consumption is calculated from the current power consumption and the polling time interval set in the software.
The energy per epoch, commonly stated in watt per hour, is a common metric and follows the calculation of \cref{energy_equ}.
 \begin{equation}
 \label{energy_equ}
    E_{k} = \frac{1}{n} \sum_{i=0}^{n} \text{{power}}(k, n) \cdot \frac{\text{{time}}(k)}{3600}, \quad \forall \, k \in [0, T]
\end{equation}

Works like Carbontracker \cite{anthony2020carbontracker}, eco2Ai \cite{budennyy2022eco2ai} and Green Algorithms \cite{lannelongue2021green} utilize this approach to gather data from the hardware.
To handle missing elements in the calculations, software like Carbontracker multiplies its results with an efficiency constant to incorporate untracked secondary power needs and efficiency losses.
With an extended focus on user experience, projects like Cloud Carbon \cite{cloudCarbon} or CodeCarbon \cite{CodeCarbon} extend the gathered knowledge and present it in analytic-based dashboards. 
Based on the calculated energy consumption and the user’s location, the average local energy mix from fossil and renewable energy sources is utilized to estimate the carbon emissions in kilograms or even tons. \cite{lacoste2019quantifying} 
On top of that, since the carbon emissions are difficult to visualize or imagine, the conversion into kilometers driven by car, flights with a plane, or the number of phones charged is a standard practice to make the user aware of the generated carbon emissions amount.

\section{Energy-Aware Training}

A major goal of this work is to generate energy awareness within the training process to finalize a well-performing model trained through hardware operated in an energy-efficient state.
Instead of compromising between performance and sustainability, we envision SM$^2$ as a strategy to prioritize two objectives at the same time.
During our initial tests, we established the energy expenditure per trained epoch as a suitable metric to calculate the efficiency of the current setup.
As discussed in \cref{energy_sec}, a precise solution for tracking the total energy consumption of all involved hardware components is still missing.
Consequently, Our work focuses on the energy efficiency of the GPU, the primary energy consumer for training deep-learning models in a parallelized and hardware-accelerated environment \cite{mittal2014survey}.

To properly monitor the energy consumption of the GPU, we decided to utilize the Carbontracker library \cite{anthony2020carbontracker}.
Due to its technical and data-driven approach compared to the other available solutions, it provides an appropriate foundation to consciously track the energy demand of hardware.
The library operates as a background service, tracking the GPU power through the hardware's System Management Interface (SMI) over time to calculate the consumed energy.
In our implementation, based on version 1.2.5. of Carbontracker, we extended its capabilities to facilitate live tracking of energy consumption instead of logging the data to files.
Throughout a callback function, the current power consumption and the energy of the current epoch can be obtained.
This integration ensures that SM$^2$ directly benefits from the relevant information within the training loop.

Since different manufacturers offer their unique interface solutions for measuring and they may differ in terms of true and measured GPU wattage, we decided to test SM$^2$ on CUDA-supported GPUs to provide reproducibility and comparability of results.
Currently, SMI provides information only about the total power consumption of the GPU, neglecting the number of processes or threads concurrently running in parallel on the GPU.
Therefore, Carbontracker is unable to partition the overall energy consumption to multiple processes, whereas our implementation of SM$^2$ is based on a sequential architecture to ensure proper tracking within these limitations.

In theory, a more comprehensive implementation could include extended hardware information logging for different manufacturers or other hardware components, such as the Central Processing Unit (CPU) or the PSU. However, the inclusion of such components would necessitate careful validation to ensure the validity of cross-comparable measurements. 
To the best of our knowledge, including the discussed restrictions, our custom implementation of Carbontracker to forward the power and energy information from the SMI directly into the training loop is currently the first and most promising solution for SM$^2$.

\section{SM$^2$: Spend More to Save More}

HPO algorithms usually form an additional layer throughout a library or framework on top of the traditional training to optimize selected hyperparameters \cite{bergstra2013hyperopt}.
With the aim to maximize the final model's performance, they usually do not account for the waste of resources to explore the hyperparameter space.
As a result, HPO is often connected with time-consuming and energy-intensive training of weak models, especially in terms of rudimentary algorithms like random or grid search.
Each setup of hyperparameters is trained, and the best version is taken as the final model.
The unpleasant side, which is usually not presented in publications or advertisements, contains the number of less-performing models, trained just for the purpose of exploring the hyperparameter space.
Together with the growing size of datasets and the complexity of machine learning models, the wasted energy adds up and further supports climate change if fossil energy resources are utilized.

As already introduced in \cref{hyp_opt}, SHA and its variations like ASHA are promising solutions and are considered state-of-the-art to the best of our knowledge.
To summarize, SHA is an iterative hyperparameter optimization method in which the number of inspected hyperparameter configurations is pruned exponentially over time until the final, best-performing model remains.
The idea of terminating less-performing models early in the training process to decrease the waste of resources supports our idea for sustainable HPO, wherefore we decided to implement SHA as a general basis for "Spend More to Save More".
As a motivation, we argue that spending slightly more energy for a single training run is more sustainable than training models multiple times, considering that such a strategy can find the optimal hyperparameter configuration satisfactory.

Based on previous research \cite{geissler2024power}, batch size and learning rate rule the energy efficiency for training a  machine learning model, whereas we focus on those two hyperparameters in this work.
It is noteworthy that SM$^2$ is not limited in this regard, however, there is not such a strong influence on efficiency by other hyperparameters.
Operating independently of the core training process, custom optimizers, models, and loss functions can still be utilized and provide evidence for the universality of our approach.

Identifying a suitable batch size lets the GPU run in an efficient power state based on the GPU utilization.
For SM$^2$ we do not utilize the wattage as a metric, the energy per epoch gathered from Carbontracker serves as an indicator of how efficiently the current batch size matches the hardware power state.
On the other hand, operating the GPU in its efficiency window may not contribute to the overall efficiency of the training run if too many epochs are necessary to complete the training.
Therefore, it is essential to work out a well-performing learning rate to speed up the model's convergence.
To solve this, we implement cyclical learning rate exploration within the SHA algorithm to identify the optimal setting \cite{smith2017cyclical}.

To generate a more efficient HPO algorithm compared to existing approaches, we split up the training into an alternating process of exploratory training and thorough training.
With exploratory training on fewer batches of the dataset and only one epoch duration, SM$^2$ identifies trends of the current selection with less energy waste to narrow down the hyperparameter space.
The thorough training represents the traditional training on the full dataset across multiple epochs.

An objective function finally unites the introduced sustainability considerations with the main aim of the model to maximize its prediction performance.
The following sections explain the main components in more detail to manifest their connection within SM$^2$.

\subsection{Batch Size Optimization}

Next to the dataset and model architecture, batch size is a crucial parameter to control the utilization of the GPU \cite{you2023zeus}.
To efficiently utilize the GPU, one might think that maximizing the GPU utilization results in less energy consumption due to the improved training speed.
However, depending on the specific hardware, the efficiency window of a GPU usually lies below the maximum utilization due to bottlenecks like data transfer or cooling issues \cite{you2023zeus}. 
As such information is not available, calculating the optimal batch size for efficiency beforehand is impossible, wherefore we utilize SHA and the energy per epoch metric as a solution.
Throughout the training, our energy-aware implementation tracks the average power and duration of each epoch and calculates the energy consumption for training one epoch.
Based on such metrics, energy awareness can be integrated into the training process.
Within the exploratory training, the energy per epoch for each batch size configuration is monitored and taken into account with the objective of removing inefficiently trained configurations.

\subsection{Learning Rate Optimization}

In addition to measuring the energy per epoch to train the model in an efficient environment, the number of epochs trained needs to be minimized to keep the overall energy consumption low.
To evaluate the learning rate effectiveness, cyclical learning rate scheduling is embedded into the exploratory training phase \cite{smith2017cyclical}.
For each batch in the exploration phase being passed through the model, a different learning rate is tested through alternation and the respective loss is documented.
As a general rule, low learning rates tend to unnecessarily extend the training process, whereas larger learning rates may increase loss fluctuations \cite{smith2018disciplined}.

\begin{figure}[!t]
 \centering
 \includegraphics[width=\linewidth]{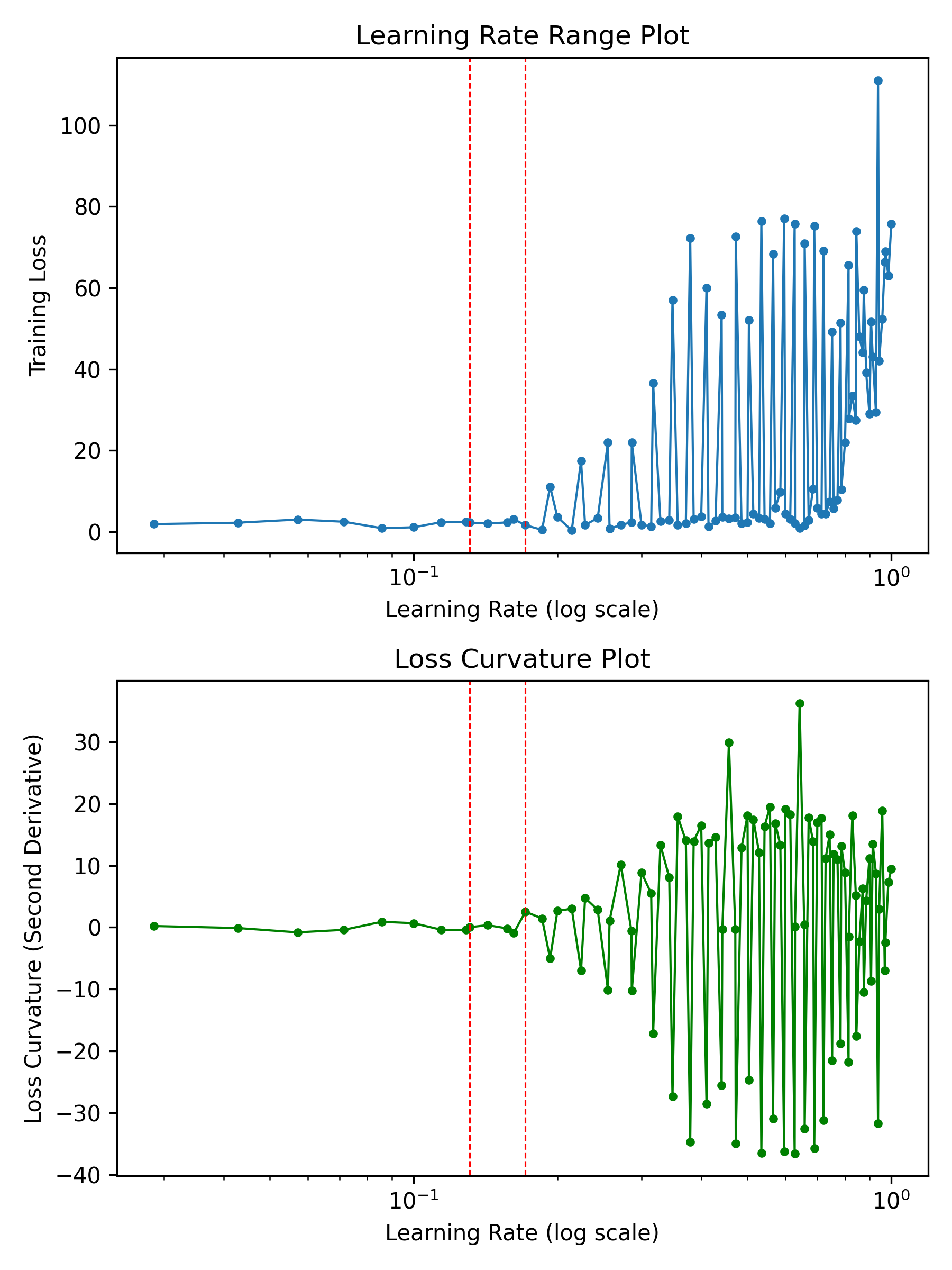}
 \caption{Computational analysis of loss curvature through the sliding window to select the area with the largest stable learning rate.}
 \label{lr_curvature}
\end{figure}

As shown in \cref{lr_curvature}, we analyze the inspected learning rates computationally by calculating the second derivative to receive the curvature of the loss.
Through a sliding window approach, we can identify windows with a low curvature.
By sorting the windows based on the curvature and the mean learning rate of the window, we can identify the largest learning rate that maintains stability throughout the training process.
In the case of \cref{lr_curvature}, we marked the selected learning rate window with two dotted lines.
Selecting the largest learning rate within the identified window further maximizes the learning rate selection in terms of shrinking the training time.

\subsection{Objective Function}
\label{obj_fnc}
To include the individual optimization targets in the training process, we introduce an objective function that determines the best-performing configurations of the current set during the exploratory phase.
Following the general SHA rule, the objective is necessary to halve the number of configurations until only one configuration remains.
Further, since SM$^2$ envisions the hyperparameter optimization as a holistic system, especially from the sustainability aspect, it is a key necessity to merge the observations for each hyperparameter to finalize a combination that fulfills the needs.

The introduced objective function in \cref{objective} accounts for the performance metric of the trained model, the consumed energy per epoch, and the selected, stable learning rate.
To compare those attributes, we rescale each list of attributes into the range between 0 and 1 across all explored configurations.
Thus, the calculated objectives as well as the three attributes themselves are independent of any influences from the specific model and the utilized hardware.
Additionally, depending on the nature of the attributes, we invert them to either reward higher values, like for the learning rate, or lower values, for instance, the energy consumption. 
Two parameters $\alpha$ and $\beta$ allow the balance between the three attributes.
They need to be set beforehand as described in the following \cref{exp_setup}.
For each configuration, the objective is calculated whereas after a full exploratory iteration, the less-performing half is dropped.

\begin{equation}
\label{objective}
\begin{aligned}
    \text{f}(\alpha, \beta) = &\alpha \times \text{P} + (1 - \alpha) \times (\beta \times \text{E} + (1 - \beta) \times \text{LR}) \\[1ex]
    \text{P} = &\text{ Performance;} \\
    \text{E} = &\text{ Energy;} \\
    \text{LR} = &\text{ Learning Rate}
\end{aligned}
\end{equation}

Once there is only one configuration remaining, the SM$^2$ algorithm remains in thorough training mode, with only a focus on further learning rate optimization.
From this point onwards, it can be assumed that the best possible efficiency per epoch has been found for the training and hardware combination and that it is now only possible to optimize the training duration to minimize energy demand.

\subsection{Algorithm}

The SM$^2$ algorithm follows the nature of the introduced SHA approach, implemented in a sequential manner due to the energy consumption tracking limitations.
As shown in \cref{alg:algorithm}, SM$^2$ consists of multiple loops to handle configurations sequentially.
The outermost while loop represents the stopping criteria, which can be set depending on the project and the user's preferences.
Afterward, the whole training is split up into two training modes, exploratory and thorough training.
While thorough training represents traditional training, the logic and the resulting adaptions to the code of SM$^2$ happen in the exploratory mode. 
Iterating through each configuration and the number of epochs to train, exploratory training has the sole benefit of improving training based on the objective function.
Both modes are isolated from each other by backing up the models after each iteration, wherefore training and testing in exploratory mode does not contribute to the final model's performance.
On the other hand, the independence between SM$^2$ and the thorough training mode enables custom implementations by the user, for instance, to integrate the desired optimizer or early stopping metric.
Even though exploratory training contributes towards increased energy consumption, the benefit of exploring batch size and learning rate outperforms the energy consumption.
This strategy again manifests the idea of "Spend More to Save More".

\begin{algorithm}[tb]
    \caption{SM$^2$ Algorithm}
    \label{alg:algorithm}
    \begin{algorithmic}[1] 
        \STATE Initialize SM$^2$ Setup
        \WHILE{not stop}
            \FOR{mode \textbf{in} [expl, thorough]} 
                \FOR{config \textbf{in} Configs}
                \STATE Prepare config-related data
                    \FOR{epoch \textbf{in} Epochs} 
                        \IF{mode=expl}
                            \STATE Exploratory Training
                            \STATE Isolated Environment
                            \STATE Dataset Partition
                        \ELSE
                            \STATE Thorough Training
                            \STATE Full Dataset
                        \ENDIF
                    \ENDFOR
                \ENDFOR
                \IF{mode=expl}
                \STATE Evaluate Exploration
                \STATE Drop less performing configs
                \ENDIF
            \ENDFOR
        \STATE Update stopping condition
        \ENDWHILE
        \STATE \textbf{return} model, energy
    \end{algorithmic}
\end{algorithm}

\section{Experiments}
\label{experiments}
\subsection{Setup}
\label{exp_setup}
To validate the universality and effectiveness of the proposed SM$^2$ approach, we conducted experiments on three different machine learning scenarios, each utilizing a distinct model architecture and dataset.
Starting with a ResNet-18 model trained on the CIFAR-10 dataset as an initial validation of our approach, we further added an LSTM model trained on the Energy-Household dataset and a Transformer model trained on WikiText2 to check the usefulness of SM$^2$ for different complexities.
The model architectures and their respective hyperparameters were selected following best practices for each scenario. 
Throughout all experiments, model architectures and dataset preprocessing procedures were maintained consistently.
Additionally, all experiments were run three times on different hardware configurations using Nvidia RTX A6000, Nvidia Tesla V100-32GB, and Nvidia GeForce RTX 3090, ensuring hardware independence in the evaluation.

From the implementation side, we utilized PyTorch \cite{NEURIPS2019_9015} as the basic framework to thoroughly execute SM$^2$ in the hardware-accelerated GPU environment.
To isolate the energy consumption and overall performance, we added an extended initialization process.
Each spawned training setup was initialized with its own model, optimizer, and energy tracker instances.
By deep-copying the respective instances and fixing the random seed in PyTorch, we ensured that each configuration could be sequentially executed throughout the same training loop.
That being said, the only connection between the configuration is the introduced objective function and the shared dataset.

To minimize energy loss and idling of the GPU due to data transfer, we implemented a flexible batch-size system through a custom data loader.
Instead of commonly passing each batch with its respective size to the device depending on the current configuration, we preprocessed the data and preloaded batches into the GPU memory.
The preprocessed dataset was divided into batches of the smallest configuration.
With a custom iterator, configurations of larger batch sizes were trained on a temporary concatenation of batch sizes within the GPU memory.
If the size of the GPU memory cannot retain the full dataset, we utilize a queuing strategy based on a first-in-first-out strategy.

Throughout our initial tests, we experimented with the $\alpha$ and $\beta$ parameters of the objective function to check if they require any dynamic adaptions depending on the models.
Especially for the $\alpha$ value, a balance between performance metrics and sustainability is crucial to consider.
It is up to the user to decide how much weight the user wants to put into the energy consumption.
However, $\alpha$ parameter value below 0.5 usually tends to prioritize efficiency, whereas the model performance drastically decreases.
To strike a balance for including both objectives while allowing performance to keep a benefit against the efficiency, an $\alpha$ of 0.75 was found to yield the most promising results.
Since multiple configurations may achieve the desired performance, such a strategy settles the most efficient configuration out of the best-performing models.
For $\beta$, controlling the balance between batch size and learning rate schedule, we kept an equal setting of 0.5 since both have an equal possibility to improve efficiency.
All the following experiments were therefore conducted with parameters $\alpha$=0.75 and $\beta$=0.5.

Since SM$^2$ is independent of the model architecture or optimizer, users are open to customize and adapt the setup to their desired project.
The range and number of batch sizes and learning rates need to be set beforehand next to the number of iterations through exploration and thorough epochs.
For our experiments, we set up 8 different batch-size configurations with 20 inspected learning rates in the range of 0.001 to 1 during exploratory training.
The exploratory training phase was set to 1 epoch trained only on a quarter of the dataset, whereas the thorough training was set to 5 epochs and extended to 10 epochs after the final configuration was determined.

\subsection{SM\protect$^2$ Evaluation}

\begin{figure*}[!t]
 \centering
 \includegraphics[width=\textwidth]{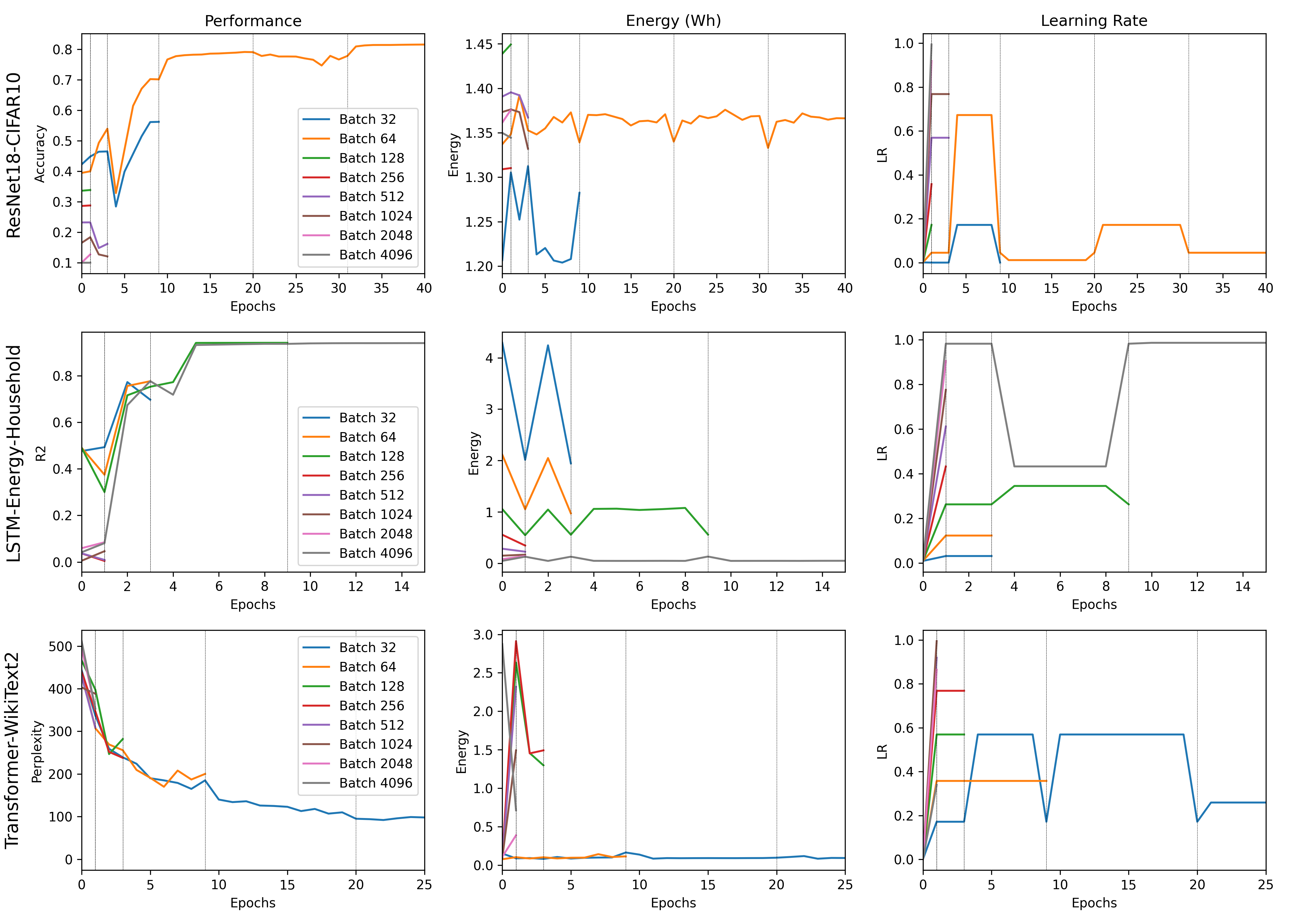}
 \caption{Evaluation of SM$^2$: Each row represents a different model and dataset combination; Columns represent Performance, Energy, and Learning Rate; Vertical Lines in the graph highlight exploratory epochs; Experiments were conducted on Nvidia RTX A6000.}
 \label{sm2_eval}
 \vspace{-5mm}
\end{figure*}
The results for evaluating the SM$^2$ approach are visualized in \cref{sm2_eval}.
Each row represents a different scenario, with the ResNet model constituting a medium complexity task, LSTM a low complexity task, and the Transformer model a high complexity respectively.
Each column plots the three acquired attributes, performance, energy, and learning rate, which are merged into the objective function as introduced in \cref{obj_fnc}.
Vertical dotted lines highlight the exploratory epochs, whereas the rest were conducted in thorough training mode.
It is worth mentioning that each plot visualizes the information gathered before normalization across configurations.
Further, we present the results for training all scenarios on the Nvidia RTX A6000.
Across our experiments, the other two hardware setups shared similar results regarding the objective function decision process while operating on different energy efficiency levels.
Due to the restriction on CUDA-based hardware supporting the SMI, we have not been able to test our approach on differing setups from other manufacturers.

Starting with the performance, we visualize the common metrics for each scenario.
For the accuracy and R$^2$ score of the first two plots, higher values indicate increasing model performance, whereas, for the last scenario, lower perplexity values state higher performance.
Across all three scenarios, the identified, final configuration selected through successive halving was able to converge the training process.

However, a great explanation can be drawn from the charts when considering the objective function.
The large initial range of accuracy in the ResNet scenario from 0.1 for the weakest up to 0.45 for the best-performing scenario rules the decision by passing the best-performing models to the next round.
Due to the small differences in energy consumption across the configurations, the slightly higher energy consumption for Batch 64 is tolerated due to the $\alpha=0.75$ passing more importance to the performance.
On the contrary, even though the $R^2$ range for the second scenario shares similarities to the first scenario, the significantly larger range of the tracked energy, especially due to poor performing smaller batch sizes, leads to an increased impact within the objective function.
As a decision of SM$^2$, training of batch 4096 is continued due to the comparable low energy per epoch and finally outperforms the remaining configurations.
For the third, transformer scenario, perplexity does not influence the beginning of the training. 
Similar to the second scenario, the energy attribute rules the decision towards the smaller, more efficiently performing configurations.

Since energy and learning rate contribute to the final objective throughout the same ratio set with $\beta=0.5$, also the learning rate, shown in the third column, contributes to the decision process.
Even though the selection through the cyclical learning rate process seems coincidental during the first exploratory phase, larger learning rate selections usually support the final configuration decision within later epochs.
Therefore, we assess the learning rate as a decision supporter towards the advanced epochs with fewer configurations remaining.

To summarize, for large deviations within the performance, SM$^2$ neglects efficiency to a certain extent.  
As a matter of fact, training a model solely based on sustainability aspects may result in another training run with additional energy consumption due to unsatisfactory performance.
On the other hand, SM$^2$ offers great potential to isolate efficient configurations if similar performance is present.

\subsection{Energy Compensation}
\begin{table}[!t]
    \centering
    \begin{tabular}{lrrrrr}
        \toprule
        Experiment & $\alpha = 1$ & $\alpha = 0.75$ & vanilla & parity\\
        \midrule
        ResNet18    & 49.7      & 45.8 (-8\%)       & 26.0      & 1.76 \\
        LSTM        & 28.0      & 14.8 (-47\%)      & 16.4      & 1.11   \\
        Transformer & 48.5    & 40.7(-16\%)       & 69.4      & 1.70   \\
        \bottomrule
    \end{tabular}
    \caption{Total energy demand (Wh), the reduction between $\alpha=1.0$ and $\alpha=0.75$ (\%), and the parity to compensate the energy overhead compared to vanilla training}
    \label{tab:totalenergy}
    \vspace{-5mm}
\end{table}

\cref{tab:totalenergy} presents the total energy demand in watt-hours (Wh) for each experiment, demonstrating the reduction in energy demand between $\alpha=1.0$ and $\alpha=0.75$, along with the parity factor. 
To analyze the results in terms of their sustainability, we executed the scenarios with $\alpha=1.0$ and a vanilla training without the SM$^2$ approach.
The $\alpha=1.0$ setting signifies a scenario where the objective function only considers the model performance, whereas the energy efficiency improvement can be calculated across the measurements.
The reduction percentage provides insights into how much energy consumption can be saved by incorporating sustainability aspects into the objective function. 
Parity represents the number of manual HPO explorations needed to compensate for the energy investment in SM$^2$ compared to vanilla training.

As shown in \cref{tab:totalenergy}, SM$^2$ managed to reduce the energy demand across all experiments by incorporating the sustainability aspects into the objective function.
For the second scenario (LSTM), the decrease in energy consumption of 47\% reflects the results depicted in \cref{sm2_eval}, where the energy-based selection of the larger batch size played a crucial role.
In terms of final model performance, the first two scenarios shared the same final performance, while for the third scenario (Transformer), the perplexity was reduced by around 15 percent points when setting $\alpha=1.0$.

Compared to the vanilla experiment, trained with a traditional training loop while keeping identical parameter setups and early stopping metrics, each setup requires at minimum two manual HPO explorations for compensation.
An additional run with a doubled amount of 16 configurations and 40 inspected learning rates resulted in parity between 5 and 6 across the scenarios.
Therefore, the parity factor serves as a useful metric to estimate the energy demand and provides evidence for the "Spend More to Save More" concept, emphasizing the reduction of energy waste through inefficient HPO.

\section{Conclusion and Future Work}
In this paper, we presented "Spend More to Save More (SM$^2$)", an energy-aware implementation for sustainable hyperparameter optimization.
Our approach enables the simultaneous optimization of both model performance and efficiency, addressing the critical need for energy-aware machine learning practices.
We presented an objective function as the core of SM$^2$ to beneficially balance model performance and energy consumption.
Although we give more weight to performance, energy efficiency can be the deciding factor, especially in the area of equally performing configurations, to optimize the training process as efficient as possible.
Our experiments prove the idea of decreasing the energy demand without significant performance loss, especially when considering the whole development life cycle.

As a part of our future work, we envision an improved version of SM$^2$ to include more energy information from other hardware components and manufacturers.
We envision a solution that dynamically adapts to the given hardware setup and fluently scales to parallelized and multi-GPU environments.
Extended work should also adapt to less prominent hyperparameters to further maximize energy efficiency.
As an outlook, SM$^2$ can be provided in the form of a library to be integrated to the traditional training workflows, offering a practical solution for researchers seeking to incorporate energy-awareness into their machine learning models. 
\newpage

\bibliographystyle{named}
\bibliography{ijcai24}

\end{document}